\documentclass[letterpaper]{article} 
\usepackage{aaai2026} 
\usepackage{times}  
\usepackage{helvet}  
\usepackage{courier}  
\usepackage[hyphens]{url}  
\usepackage{graphicx} 
\usepackage{natbib}  
\usepackage{caption} 
\usepackage{algorithm}
\usepackage{algorithmic}

\usepackage{amsmath}
\usepackage{enumitem}
\usepackage{booktabs}
\usepackage{multirow}
\usepackage[table]{xcolor}
\usepackage{soul}

\usepackage[most]{tcolorbox}
\definecolor{bggray}{rgb}{0.95, 0.95, 0.95}
\usepackage[%
    framemethod=tikz,
    skipbelow=\topskip,
    skipabove=\topskip
]{mdframed}
\mdfsetup{%
    leftmargin=0pt,
    rightmargin=0pt,
    backgroundcolor=bggray,
    middlelinecolor=black,
    roundcorner=3
}
\newtcolorbox[list inside=prompt,auto counter,number within=section]{prompt}[1][]{
    colbacktitle=black!60,
    fonttitle=\small,
    coltitle=white,
    fontupper=\footnotesize,
    boxsep=4pt,
    left=0pt,
    top=0pt,
    bottom=0pt,
    boxrule=1pt,
    #1,
}

\usepackage{fancyvrb}
\usepackage{fvextra}
\usepackage{verbatimbox}

\DefineVerbatimEnvironment{MyVerbatim}{Verbatim}
  {frame=single,fontsize=\small,breaklines=true,breaksymbolleft={} }

\usepackage{xcolor}
\newcommand{\answerYes}[1]{\textcolor{blue}{#1}} 
 
\newcommand{\answerNA}[1]{\textcolor{gray}{#1}}

\usepackage{newfloat}
\usepackage{listings}
\DeclareCaptionStyle{ruled}{labelfont=normalfont,labelsep=colon,strut=off} 
\floatstyle{ruled}
\newfloat{listing}{tb}{lst}{}
\floatname{listing}{Listing}
\title{
Intergroup Bias Predicts Expressions of Social Judgment and Empathy \\ in the \textit{r/AmITheAsshole} Reddit Community}

\author {
    Yu Hou \quad
    Hal {Daum\'e III} \quad
    Rachel Rudinger
}
\affiliations {
    University of Maryland\\
    \texttt{\{houyu,hal3,rudinger\}@umd.edu}
}

\newif\ifnotes
\notesfalse 

\definecolor{blue-d}{HTML}{0118D8}
\definecolor{teal-d}{HTML}{48A6A7}
\definecolor{purple-d}{HTML}{9112BC}
\definecolor{green-d}{HTML}{3CB371}

\newcommand{\yhedit}[1]{\ifnotes{\color{purple-d}#1}\else #1\fi}
\newcommand{\yhadd}[1]{\ifnotes{\color{blue-d}#1}\else #1\fi}

\begin{document}

\maketitle

\begin{abstract}

People share and understand others' experiences differently based on their social identities; often, people empathize more when the other person belongs to their ingroup compared to an outgroup. While this intergroup empathy gap is well-studied in social psychology, it remains unclear how it manifests in naturally occurring online social interactions.
To this end, we measure how commenters express social judgments, as an indirect outcome of their understanding towards the posters, and empathy in text, based on whether commenters share the poster's gender or are of similar ages.
We focus on posts where the poster self-reports these demographics, and analyze 126,716 comments associated with 36,163 posts from 18,111 commenters in r/AmItheAsshole, while controlling for commenter and poster group identities, post features (topic, length, and score), and both commenter- and post-level variance.
Our results show that when commenters and posters are ingroup, in either gender or age, commenters are more likely to provide a Not the Asshole verdict and show greater text-based understanding as measured by LLMs. We also observe that older commenters behave differently from younger ones.
Furthermore, in a subset of posts where the gender and age of the other party in the conflict are explicitly mentioned, we find that commenter-poster ingroup effects are moderated by the group relationships between the poster and the other party when commenters make social judgments.
Our work highlights the importance of intergroup context and social group nuances in understanding online discourse.

\end{abstract}


\section{Introduction}
\label{sec:intro}

People live in social groups and define themselves through their group memberships~\cite{Tajfel1986-rr}. These social identities often drive people to behave differently towards ingroup members compared to outgroup members, where group status is defined by shared or dissimilar social identities.
Understanding intergroup perceptions helps identify stereotypes~\cite{chen-etal-2021-the-role} and is particularly relevant in online communities given the continuous growth of online social interactions.
For example, othering, the process of categorizing an outgroup as different, can occur in food reviews~\cite{Luo_Gligorić_Jurafsky_2024} and blogs~\cite{Gerard_Weninger_Lerman_2025}, causing harms;
ingroup behavior plays an important role in influencing community experiences~\cite{nettasinghe-etal-2025-in-group-love, rotman-etal-2025-beyond-moral, weld2025conversationalstructurestyleshape}.

Building upon existing research, we focus on a common form of online interaction: a poster shares a personal experience, potentially asking for advice, and a commenter replies. Such scenarios may reflect how a commenter perceives and understands a poster differently because of the perspective that comes from their social identities.
Prior work in social psychology suggests that people show more understanding toward ingroup members.
This tendency---empathizing more with ingroup relative to outgroup members, or even feeling pleasure in response to outgroup pain---is known as the ``intergroup empathy gap''~\cite{cikara-etal-2011-us-and-them, CIKARA2014110, smith-van-dijk-2018-schadenfreude}.
The empathy gap thus influences how people treat or react to each other.
However, few studies have examined these intergroup effects in large-scale, real-life online social interactions, while controlling for possible confounds.

\begin{figure}[t]
\centering
\includegraphics[width=\linewidth]{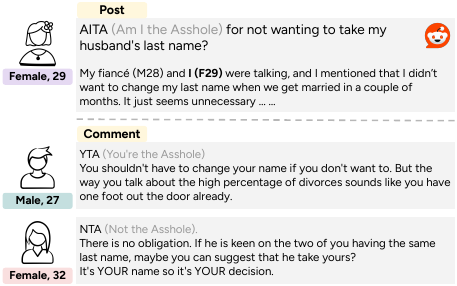}
\caption{Example post and comments from r/AmItheAsshole, including user gender and age.
We use self-reported gender and age to study ingroup behaviors. In this example, the poster is a 29-year-old female in conflict with her fianc\'e (the other party; a 28-year-old male). The commenters include a 27-year-old male and a 32-year-old female.
}
\label{fig:example}
\end{figure}


\paragraph{Present work.} In this work, we analyze social scenarios from r/AmItheAsshole (AITA) posts that explicitly mention gender and age (as shown in Figure~\ref{fig:example}).\footnote{r/AmItheAsshole is a subreddit community where people post stories about interpersonal conflicts, asking other users to judge if they are the ones to blame; for more details see \S\ref{sec:data}.}
We focus on gender and age social groups as they are not only easier to extract but also less variable compared to race/ethnicity and religion~\cite{fiske-2017-prejudices-in-cultural}.
\yhedit{We hypothesize that the intergroup relationship between the commenter and the poster plays a role in the commenter's expressed understanding of the poster's experience. This understanding is then manifested indirectly in the downstream social judgments: NTA (Not the Asshole; the poster shouldn't be blamed) or YTA (You're the Asshole; the poster should be blamed) verdicts, or more directly in their open-ended language expressing empathy.\footnote{This kind of understanding expressed through the comments is a proxy for cognitive empathy~\cite{BOCKAROVA201647}. \yhadd{We don't imply that judgments necessarily reflect understanding; rather, empathy is one of possible factors influencing the final verdict.}}}

Informed by social theories regarding intergroup bias in gender~\cite{dion-1979-status-equity, Boldizar1988, vonk-olde-monnikhof-1998-gender-subgroups} and asymmetric age intergroup bias~\cite{chasteen-2005-seeing-eye, cary-etal-2013-does-age, blanke-etal-2015-nice-to}, our research questions are formulated as follows:

\begin{itemize}[leftmargin=0pt]
    \item[] \textbf{RQ1: Ingroup effects in social judgments.} When sharing the same gender or a similar age, is the commenter more likely to give an NTA verdict? Do intergroup biases operate similarly for younger and older commenters?
    \item[] \textbf{RQ2: Ingroup effects in expressed text-based empathy.} How do open-ended \textit{comments} written in gender or age ingroup conditions differ from those in outgroup conditions?
    \item[] \textbf{RQ3: Ingroup effects moderated by social contexts.} How do the gender and age intergroup dynamics between the poster and the other party in the \textit{post} influence the ingroup effects between the poster and commenter?
\end{itemize}

To answer these questions, we start by extracting user self-reported gender and age from 456,200 AITA posts and linking these to the commenters.
We leverage LLMs to carefully examine the data and predict post and comment features.
As a result, we include 126,716 comments from 36,163 posts and 18,111 unique commenters for the logistic mixed-effects analysis, controlling for poster and commenter identities (gender and age) and post features (topic, length and score), as well as considering post and commenter group-level variance. Both gender ingroup and age distance are modeled as separate variables in the regression, with an additional interaction term between commenter age and age distance to capture the uniqueness of intergenerational behaviors.
For \textbf{RQ1}, the outcome is the social judgment (NTA/YTA), whereas for \textbf{RQ2}, the outcomes are predicted text-based empathy dimensions in emotional reactions and interpretations, adapted from~\citet{sharma-etal-2020-computational}.
For \textbf{RQ3}, we focus on a subset of data with {71,887} comments on {20,464} posts where the gender and age of both the original poster (OP) and the other party in the conflict are explicitly mentioned.
This allows us to include the other party's gender and age in the regression to observe the fine-grained ingroup effects between commenters and OPs.

\paragraph{Summary of findings.} Our analysis indicates that, on average, commenters are 3.18 \textit{pp} more likely to give an NTA verdict when they share the same gender as the poster; similarly, for every unit reduction in the age gap (meaning the age ratio narrows by a factor of 1.5), commenters are 1.03 \textit{pp} more likely to give an NTA verdict, an effect that is stronger among younger commenters \textbf{(RQ1)}.
On the empathy dimensions, we find that commenters are 2.38 \textit{pp} more likely to express emotional reactions and 1.66 \textit{pp} more likely to convey interpretations when they share the poster's gender. We observe a similar pattern for age: as the age ratio narrows by a factor of 1.5, commenters are 1.16 \textit{pp} more likely to express emotional reactions and 0.88 \textit{pp} more likely to convey interpretations, with these effects being stronger among relatively older commenters \textbf{(RQ2)}.
Finally, our results suggest that when commenters share the same gender as the poster but not the other party in the post, they are 5.4 \textit{pp} more likely to give an NTA verdict towards the original poster; they are also 1.53 \textit{pp} more likely to give NTA per unit reduction in the commenter–poster age gap, when the poster and the other party have a large age gap \textbf{(RQ3)}.


\paragraph{Implications.} Overall, we find correlations between gender and age intergroup bias and commenters' social judgments and open-ended comments in the AITA community.
This highlights that social media serves as a critical lens for understanding societal dynamics~\cite{chew_predicting_2021}.
Moreover, given the increasing use and significant influence of AI in online communication~\cite{Ruffin_Seo_Xiong_Wang_2024, hou-etal-2025-language, Glickman2025, møller2025impactgenerativeaisocial}, it is important to consider the role of intergroup bias, whether originating in human conversations or learned by AI, when AI takes on different roles in a community, such as a poster, a commenter, or a mediator.

\section{Background and Related Work}
\label{sec:background}

\paragraph{Studies on the r/AmItheAsshole community.} Data from the AITA subreddit has been widely studied across various domains.
Research treats AITA posts as moral dilemmas to identify their topics~\cite{Nguyen_Lyall_Tran_Shin_Carroll_Klein_Xie_2022} and moral dimensions~\cite{nguyen_chen_carroll_tran_klein_xie_2024}, as well as to evaluate LLMs on morality~\cite{sachdeva_tom_normative_2025} or social sycophancy~\cite{cheng2025elephantmeasuringunderstandingsocial}, or study social norms~\cite{forbes-etal-2020-social, de-candia-etal-2022-social-norms, vijjini-etal-2024-socialgaze}.
Regarding user interactions, comments and verdicts are analyzed to understand how moral judgments are formed~\cite{lourie_le_bras_choi_2021, efstathiadis2022explainablepatternsdistinctionprediction, botzer_analysis_of_moral_2023, Xi_Singh_2024, Betti2025, xi_moral_sparks_2025} and how blame is assigned~\cite{zhou-etal-2021-assessing, xi_the_blame_game_2024}. 
Finally, AITA data is also used to understand social narratives and norms~\cite{Giorgi_Zhao_Feng_Martin_2023, henderson-etal-2026-examining}.

\paragraph{Role of social context in language.} The social perspective in language is an important area of study~\cite{sap-etal-2020-social, zhou-etal-2023-cobra, shen-etal-2025-words}.
\citet{yang-etal-2025-socially} call for social awareness when building NLP models or developing applications based on language technologies.
These social factors include the context of the social environment, such as people's perspectives and intergroup or interpersonal relationships~\cite{Althoff_Danescu-Niculescu-Mizil_Jurafsky_2014, buechel-hahn-2017-readers, sepahpour_fard_how_do_mothers_2022, govindarajan-etal-2023-counterfactual}.
Social media text encodes people's attitudes towards different genders~\cite{voigt-etal-2018-rtgender, ch-wang-jurgens-2021-using}, elicits empathy in personal narratives~\cite{shen-etal-2023-modeling, shen-etal-2024-heart}, and reveals how people empower each other~\cite{njoo-etal-2023-talkup} or express condescension~\cite{wang-potts-2019-talkdown}.
Together, these studies highlight the importance of carefully understanding human behaviors in language.

\paragraph{Relation between present and prior work.} Following prior work, we view AITA posts as real-life social scenarios that people encounter. 
When users read these narratives, they comment based on their own judgments and personal experiences, related to their social roles.
We posit that the language used in these social interactions implicitly reflects ingroup bias---specifically, by showing more understanding towards posters who self-report the same gender or similar age.
While prior AITA research has studied social factors including gender and age, to the best of our knowledge, we are the first to study the intergroup effects at scale by considering the relationship between the poster and the commenter, while carefully controlling for other factors.


\section{Materials and Methods}
\label{sec:data}

\subsection{Data and Data Processing}

\paragraph{r/AmItheAsshole posts and comments.} The r/AmItheAsshole community is a popular English-language subreddit where users anonymously share personal dilemmas and ask the community to judge whether they are to blame in interpersonal conflicts.
Community users engage by commenting on posts or interacting via votes and replies.
According to the community guide, commenters are asked to include one of seven verdicts in their top-level comment: YTA (You're the Asshole), YWBTA (You Would Be the Asshole), NTA (Not the Asshole), YWNBTA (You Would Not Be the Asshole), ESH (Everyone Sucks Here), NAH (No Assholes Here), or INFO (Not Enough Info).
\yhedit{The original data is collected from Pushshift dumps spanning 2014 to 2022~\cite{Baumgartner_Zannettou_Keegan_Squire_Blackburn_2020}. In total, there are \underline{456,200} posts and \underline{36,951,098} top-level comments after removing the non-existing ones (empty or deleted). Comments here are possibly not associated with any of the 456,200 posts, as the parent post might have been deleted.}

\begin{figure}[t]
\centering
\includegraphics[width=\linewidth]{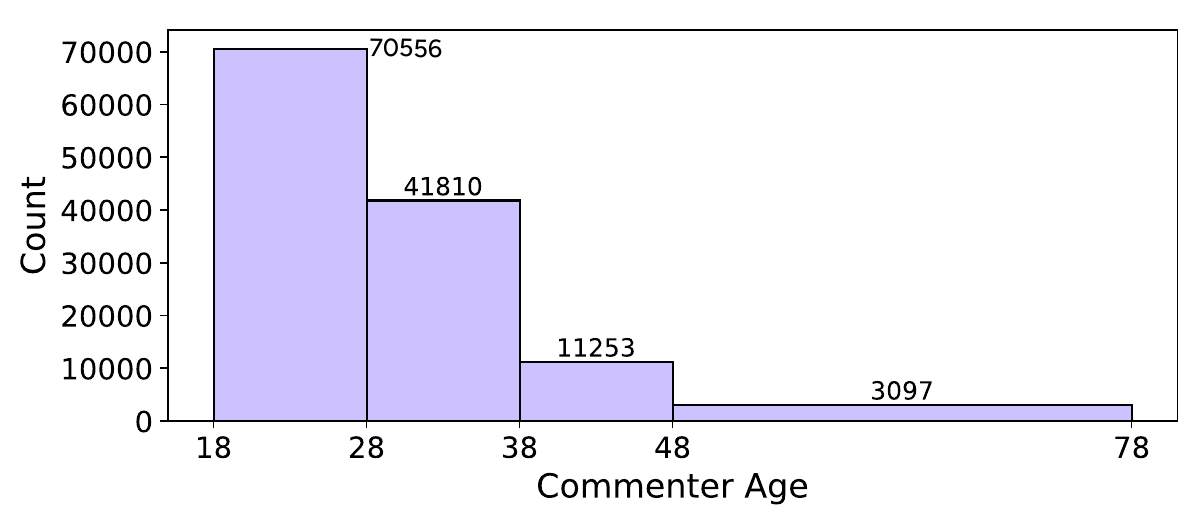} \\
\includegraphics[width=\linewidth]{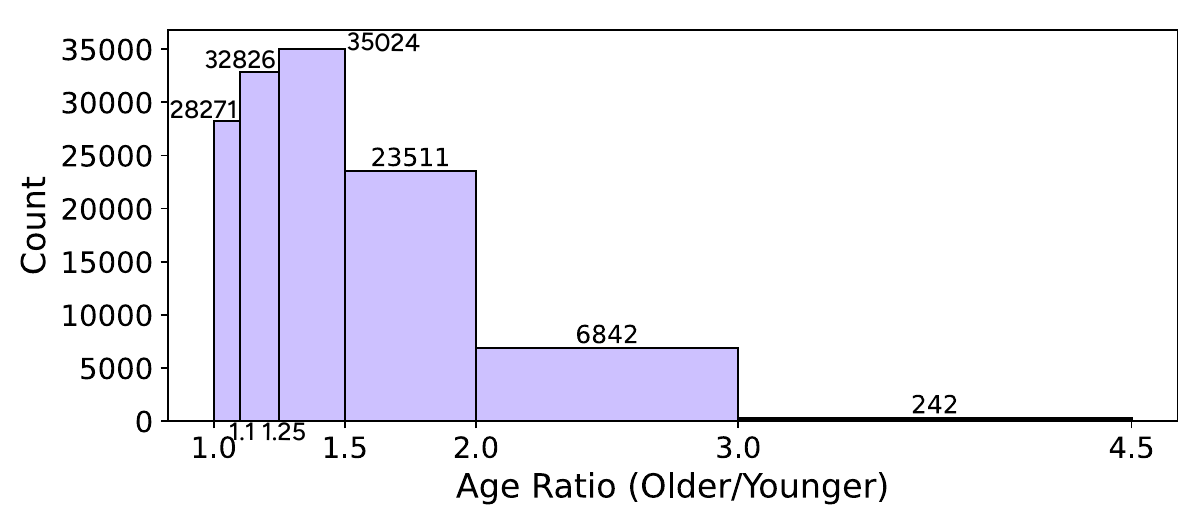}
\caption{Distribution of commenter age and age ratios for the final 126,716 comment-level interactions. We note that a large portion of users belong to younger groups and interact with others of relatively similar generations.}
\label{fig:age-dist}
\end{figure}

\paragraph{Self-reported user database.} In our setup, to study how commenters understand the post and are influenced by poster identities, we build the intergroup relationship based on a post with explicitly mentioned gender and age, and a commenter with known demographics.
Reddit user identities are anonymous by default, but there are many efforts to collect and predict users' self-disclosed attributes~\cite{beller-etal-2014-im, burger-etal-2011-discriminating, burger2006exploration, gjurkovic-etal-2021-pandora}.
In r/AITA, while users typically comment on a post without exposing their gender and age, their social identities can be approximated using their posts where they mention these details (such as ``F29'' in Figure~\ref{fig:example}), as part of the community norm when sharing personal stories.\footnote{We acknowledge the possibility that users may not self-report accurate information online for various reasons, and the limitations of only using binary genders, which dominate the data distribution.}

\yhedit{We start with regex-based gender and age extraction over the total 456,200 posts, following~\citet{de-candia-etal-2022-social-norms}. However, after manually validating the data, we find many false positives (e.g., ``I'm 24'' becomes an extraction representing male and 24 years old). Considering the limited manual fixes we can apply for this approach, we incorporate an LLM-based extraction approach to improve the accuracy of building the user database.
While arguably we can only rely on LLM extraction results, we find cases where LLMs try to predict or guess rather than extract the gender or age (e.g., ``my husband'' becomes male) or do not extract the OP's gender or age when the post involves many people.
To make sure this step doesn't introduce noisy labels, we compare the extraction results between the regex-based and LLM-based approaches, and only retain users when both gender and age are extracted and agreed upon by both approaches.}
\yhadd{Specifically, we deploy the open-weight \texttt{Gemma-3-27B} with vLLM~\cite{kwon2023efficientmemorymanagementlarge}. We then run separate asynchronous calls with a temperature of 0 for each of the 456,200 posts. Prompt details are in Appendix~\ref{app:llm-prompt} (Prompt A.1).\footnote{Huggingface model name: \url{google/gemma-3-27b-it}; a single A100 is used for this experiment.} Overall, there are \underline{119,063} posts with self-disclosed gender and age, verified by both approaches.}

Further, we calculate the year of birth based on the extracted age, accounting for the fact that users post and comment at different times.
Given that users might post multiple times with conflicting information, we only include those who report their gender consistently and have a year of birth difference of at most one year (which is likely due to calculation error).
Overall, this process yields \underline{105,376} unique users with known gender (including 72,921 females and 32,455 males) and year of birth ranging from 1935 to 2012 (median: 1998; Q1-Q3: 1994-2002).

\yhadd{To examine the selection bias from self-disclosure, we compare user activity between self-disclosed and other users. Of the 2,441,511 total users, 503,753 only post (65,346 of the 105,376 self-disclosed users only post too) and 1,720,007 only comment. Among users who both post and comment, self-disclosed users (N=40,030) are slightly more active than other users (N=177,721), with more posts (Mean=2.12, SD=2.35 vs. Mean=1.57, SD=1.29), more comments (Mean=28.85, SD=162.53 vs. Mean=26.33, SD=164.86), and moderately longer comments on average (Mean=349.63, SD=305.26 vs. Mean=312.75, SD=313.09). No strong differences are found between the two groups.}


\paragraph{Final dataset.} 
\yhadd{With the user database, we are now able to map comments to users with known gender and age, and associate comments with one of 119,063 posts where the poster self-disclosed their gender and age. This leads to a total of \underline{194,576} comments.}

\yhedit{Next, to prepare one of the key outcome variables, the final verdict, we employ a keyword-based extraction approach as in~\citet{plepi-etal-2022-unifying, welch-etal-2022-understanding}. We further aggregate the verdicts into two binary categories: (1) {NTA} (including NTA, YWNBTA, and NAH), indicating the poster is not to blame, and (2) {YTA} (including YTA, YWBTA, and ESH), indicating the poster is to blame. We exclude comments with the INFO verdict from our analysis.}
To summarize, we further process the data with the following criteria: (1) the comment is associated with a commenter with known gender and age who is at least 18 years old; (2) the comment is linked to a post containing explicitly mentioned poster gender and an age of 18 or older; (3) the comment is a top-level comment with a clearly extracted verdict falling under the aggregated NTA or YTA; (4) the commenter has at most 100 comments;\footnote{\yhadd{There are very active AITA users dominating the activities, so to focus on common user behaviors, we set the threshold as 100 according to \url{https://www.reddit.com/r/AmItheAsshole/wiki/faq/}, where 100 Top Voted Comments is Supreme Court Just-ass.}} and (5) the commenter is different from the poster. In the end, we have \underline{126,716} comments from \underline{18,111} commenters associated with \underline{36,163} posts (2018-2022).\footnote{Bot posts and posts with short content were excluded because of the self-disclosure requirements. The resulting post character lengths range from 291 to 15,199 (Mean = 2,424, SD = 954.58).}


\subsection{Feature Preparation}

\begin{table}[t]
\small
\centering
\begin{tabular}{@{}cccccc@{}}
\toprule
\textbf{Romance} & \textbf{Family} & \textbf{Friendship} & \textbf{Work} & \textbf{Public} & \textbf{Society} \\ \midrule
10,102 & 17,716 & 5,088 & 1,690 & 1,502 & 65 \\ \bottomrule
\end{tabular}
\caption{Post topic distribution across 36,163 posts. \underline{Work} is short for work/school, and \underline{Public} refers to public strangers. We merge \underline{Public} and \underline{Society} into one topic during the regression analysis, given the small predicted portion overall.}
\label{tab:post-topic}
\end{table}
\begin{table}[t]
\centering
\begin{tabular}{@{}l|cc@{}}
\toprule
\multicolumn{1}{c}{} & \textbf{Commenter: M} & \textbf{Commenter: F} \\
\midrule
\textbf{Poster: M} & \textbf{8,691} & 34,778 \\
\textbf{Poster: F} & 15,263 & \textbf{67,984} \\
\bottomrule
\end{tabular}
\caption{Counts of comment-level gender interactions (M=Male, F=Female). The diagonal represents \textbf{ingroup} cases. \yhedit{We note that female-female interactions make up the majority. To account for this, we include commenter and poster gender as controls, so the ingroup gender variable captures effects beyond baseline gender differences.}}
\label{tab:gender-dist}
\end{table}

\paragraph{Post-related features.} To study the ingroup effects on comment-level outcomes, we control for post-related features, including:
\textbf{poster gender ($P_G$)}, a binary variable indicating the self-reported gender of the poster ($1$ for male, $0$ for female); \textbf{poster age ($P_A$)}, the natural logarithm of the poster's self-reported age, $\ln(\text{age})$; \textbf{post length ($P_L$)}, the natural logarithm of the post length in characters; \yhadd{\textbf{post score ($P_S$)}, derived from the post metadata, capturing overall post engagement and z-score normalized for model fitting;} and \textbf{post topic ($P_T$)}. The natural logarithm is used to account for the skewed distribution.
As for the topic categorization, given that AITA posts are mainly about interpersonal conflicts, we define topics based on relationship categories from~\citet{de-candia-etal-2022-social-norms}. Topics include: romance, family, friendship, work/school, public strangers, and society.
\yhadd{To determine the topic for each post, we prompt \texttt{gpt-4o-mini-2024-07-18} with temperature 0 in separate calls}.\footnote{\url{https://platform.openai.com/docs/models/gpt-4o-mini}}
The detailed prompt is provided in Appendix~\ref{app:llm-prompt} (Prompt A.2).
\yhadd{We validate the predictions on 100 randomly sampled posts and find that 7 predictions are inaccurate. The errors are mainly from posts involving multiple parties with complicated relationship.}
We merge the public and society topics during the regression analysis, given that only 65 out of 36,163 posts are predicted as society (Table~\ref{tab:post-topic}).
We also include the \textbf{post ID ($P_\text{ID}$)} to account for grouping effects.

\paragraph{Commenter-related features.} We include \textbf{commenter gender ($C_G$)} and \textbf{commenter age ($C_A$)}, calculated the same as poster gender and poster age, controlling for baseline gender and age effects from commenters.
Note that both $C_G$ and $C_A$ are extracted from the commenter's past self-disclosed posts, not derived from the comment itself.
The \textbf{commenter ID ($C_\text{ID}$)} is included to control for repeated observations.

\paragraph{Intergroup features.} To capture effects related to our ingroup bias hypothesis, we define \textbf{ingroup gender ($\text{IN}_G$)} as a binary variable, where $1$ indicates $P_G = C_G$ (the commenter shares the same gender as the poster) and $0$ indicates they are in different gender groups. Given that age is a continuous variable, we define \textbf{ingroup age ($\textbf{IN}_A$)} based on the distance between $P_A$ and $C_A$ ($\ln(\text{age})$), such that $0$ implies the commenter is the same age as the poster. Specifically, to ease the interpretation, we define:
$$\textbf{IN}_A = -\frac{|P_A - C_A|}{\ln (1.5)} = - \left| \log_{1.5} (\text{age ratio}) \right|$$
meaning that a one-unit increase in $\textbf{IN}_A$ corresponds to the age ratio narrowing by a factor of 1.5.
For example, for a commenter aged 54 and a poster aged 24 (or a poster aged 54 and a commenter aged 24), $\text{IN}_A = -2$.

Table~\ref{tab:gender-dist} shows the gender distribution, while Figure~\ref{fig:age-dist} displays the age distribution.

\begin{figure}[t]
\centering
\includegraphics[width=\linewidth]{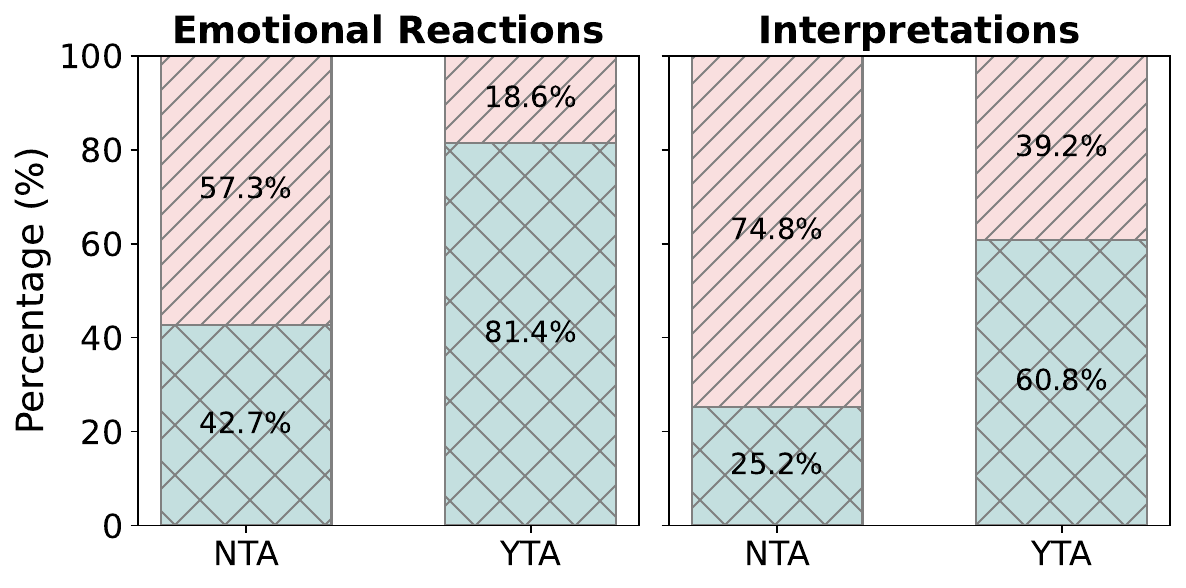}
\caption{Distribution of social judgment outcomes (NTA: N=88,977; YTA: N=37,739) across text-based empathy dimensions: emotional reactions $E$, interpretations $I$.
Top [\textbf{//}]\colorbox[HTML]{F9DFDF}{ } represents the percentage of comments predicted as $1$ on that dimension, while bottom [\texttimes]\colorbox[HTML]{C4DFDF}{} represents those predicted as $0$ (lacking expressed empathy).
YTA comments generally contain fewer emotional reactions and empathetic interpretations. The variance within both judgments (NTA or YTA) highlights the importance of analyzing comment language features in addition to final verdicts.}
\label{fig:outcome-dist}
\end{figure}

\paragraph{Outcomes.} We study two types of outcomes expressed in a comment, reflecting how the commenter perceives the poster's story, covering three dimensions:

(1) Social judgment (RQ1 and RQ3; widely discussed in past work): This is expressed by the aggregated binary \textbf{verdict ($V$)}, a variable where we encode NTA as $1$ and YTA as $0$. We hypothesize that ingroup gender or age correlates with NTA\textemdash meaning the poster is not considered blameworthy, in contrast to a YTA verdict, where the commenter might show less understanding towards the poster.

(2) Text-based empathy (RQ2): When representing a comment with only the verdict, the rich language meanings in the other parts of the comment are lost. As seen in the first example in Figure~\ref{fig:example}, even if the verdict is YTA, the commenter may still communicate understanding (e.g., ``\textit{You shouldn't have to change your name if you don't want to.}'').

Since there are no existing measures specific to AITA comments capturing text-based empathy, we develop two scales: \textbf{emotional reactions ($E$)} and \textbf{interpretations ($I$)} based on EPITOME, a framework established for understanding empathy in text-based mental health Reddit interactions~\cite{sharma-etal-2020-computational}. Recent work shows that LLMs have the potential to perform on par with humans in this subjective task~\cite{kumar2025largelanguagemodelsreliable}.
We revise the original questions as follows:
\begin{itemize}
\item \textbf{Emotional Reactions:} Does the comment express or allude to warmth, compassion, concern, or similar feelings from the commenter towards the poster?
\item \textbf{Interpretations:} Does the comment communicate an understanding of the poster's experiences and feelings?
\end{itemize}
\yhedit{Each question comes with three options (rating scale 0--2) for both human annotators and LLMs, enabling fine-grained judgments. Prompt details are in Appendix~\ref{app:llm-prompt} (Prompt A.4). During later analysis and modeling, positive ratings are merged to improve the reliability of the scales, where $1$ indicates the comment expresses emotional reactions or interpretations, while $0$ means a lack of expressed empathy.}
\yhadd{To validate the instructions, two graduate students annotate 40 randomly sampled comments, achieving Cohen's Kappa of 0.56 and 0.46 (moderate agreement) for Emotional Reactions and Interpretations, respectively. The main disagreements arise in cases where the comment expresses empathy toward other parties in the post, rather than the OP. This observation is incorporated into the LLM prompt.
Specifically, we use \texttt{gpt-4o-mini-2024-07-18} under the same setup (asynchronous single inferences) to annotate all comments.
Figure~\ref{fig:outcome-dist} shows the relationships between the predictions and social judgments.
Authors further annotate 200 randomly sampled comments.
On this test set, the precision, recall, and F1 scores for Emotional Reactions are 0.85, 0.65, and 0.73, respectively; for Interpretations, they are 0.91, 0.94, and 0.92 respectively.}


\subsection{Regression Analysis}

\paragraph{Mixed-effects model.} We use mixed-effects models to isolate specific effects while controlling for repeated observations across groups. The formula for RQ1 and RQ2 is:
\begin{equation}
    \begin{aligned}
        \nonumber \ln\left(\frac{\pi_i}{1 - \pi_i}\right) &= \beta_0 + \beta_1 \text{IN}_{Gi} + \beta_2 \text{IN}_{Ai} \\
        &\quad + \beta_3 C_{Ai} + \beta_4 (\text{IN}_{Ai} \times C_{Ai}) \\
        &\quad + \boldsymbol{\gamma}^T \mathbf{X}i  + u_{P_\text{ID}[i]} + v_{C_\text{ID}[i]}
    \end{aligned}
\end{equation}
In this equation, $\pi_i$ denotes the probability of the outcome ($V$, $E$, or $I$) being $1$. The main independent variables are ingroup gender ($\text{IN}_G$) and ingroup age ($\text{IN}_A$). We include an interaction term between commenter age ($C_A$) and ingroup age ($\text{IN}_A \times C_A$) to test the asymmetric age ingroup bias hypothesis, which suggests that older adults may exhibit different ingroup biases compared to younger adults~\cite{chasteen-2005-seeing-eye}.
The vector $\mathbf{X}_i$ contains the remaining control covariates: poster gender ($P_G$), poster age ($P_A$), post length ($P_L$), \yhadd{post score ($P_S$)}, post topic ($P_T$), and commenter gender ($C_G$).
Here, by including the gender and age of the poster and commenter separately, we effectively control for their group-specific (e.g., male or female) base rates of issuing verdicts or showing empathy.
Finally, the model includes crossed random intercepts $u_{P_\text{ID}}$ and $v_{C_\text{ID}}$ to account for unobserved variation across unique posts and commenters.\footnote{We find no multicollinearity among the main predictors (variance inflation factors $< 2$ for all non-interaction terms).}

\paragraph{Average marginal effects.} We report population-level average marginal effects~\cite{ame, marginaleffects}. Results are interpreted in percentage points (\textit{pp}), representing the change in the probability of a positive outcome, i.e.\ giving an NTA verdict ($V$) or showing text-based empathy ($E$ or $I$). 
This is the difference between groups (e.g., $\text{IN}_G = 1$ vs. $0$) or the change associated with a one-unit increase in continuous variables (e.g., $\text{IN}_A$).


\section{Results}
\label{sec:results}

\begin{figure}[t]
\centering
\includegraphics[width=\linewidth]{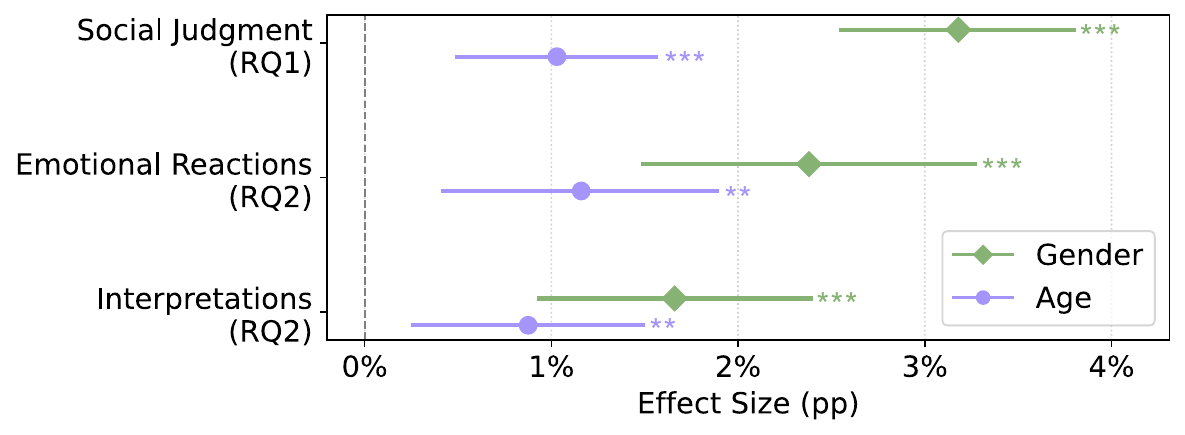}
\caption{Estimated overall gender and age ingroup effects (in percentage points) on social judgment and two text-based empathy dimensions. Positive effects indicate that when the commenter and poster share the same gender or are similar in age, commenters are more likely to give an NTA verdict or express greater empathy in their comments. 95\% confidence intervals are reported. Significance: ***$p < 0.001$, **$p < 0.01$, *$p < 0.05$. }
\label{fig:ingroup-overall}
\end{figure}

\subsection{RQ1: Ingroup Effects in Social Judgments}

Social judgments and how posters and commenters behave differently based on gender and age on AITA are widely discussed in research.
For example, \citet{Betti2025} show through a causal analysis that these judgments are not biased by gender. However, the commenters, who perceive the post from different perspectives and deliver these judgments, are often ignored in such analyses.
This is important, as these judgments are not only driven by the post content but also by the commenter's personal understanding and interpretation of the poster.
Therefore, our first set of hypotheses tests for ingroup effects in social judgments. Based on our model, we expect to observe a significant positive estimate for $\text{IN}_G$ or $\text{IN}_A$ if commenters who share the same gender or similar age as the poster give more Not the Asshole verdicts, regardless of the specific gender or age of the poster or commenter.

\paragraph{Averaged gender ingroup effects.}
Figure~\ref{fig:ingroup-overall} shows the results. In this dimension, when the intergroup relationship changes from gender outgroup to ingroup, there is a 3.18 \textit{pp} (95\% CI: [2.54, 3.81]; $p<0.001$) increase in the probability of commenters giving an NTA verdict.

\paragraph{Averaged age ingroup effects.}
On the age distance side, when the ratio between the commenter and poster narrows by a factor of 1.5 (e.g., from a commenter aged 20 reading a post by a 30-year-old---or vice versa, due to the absolute distance---to a 20-year-old commenter interacting with a poster also at age 20), commenters are 1.03 \textit{pp} (95\% CI: [0.49, 1.57]; $p<0.001$) more likely to give NTA.

\begin{figure}[t]
\centering
\includegraphics[width=\linewidth]{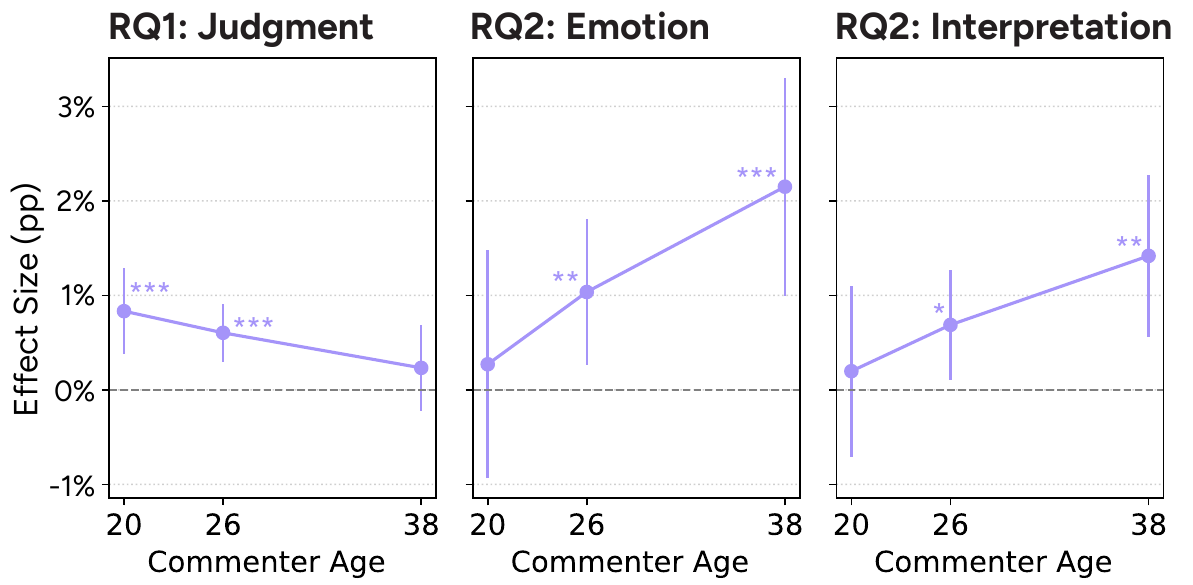}
\caption{Estimated age ingroup effects on social judgment and two text-based empathy dimensions (Emotional Reactions and Interpretations) for commenters aged 20, 26, and 38. Results indicate that older commenters exhibit weaker ingroup bias in social judgments but stronger ingroup bias in expressed text-based empathy compared to younger commenters. 95\% confidence intervals are reported. Significance notation: ***$p < 0.001$, **$p < 0.01$, *$p < 0.05$.}
\label{fig:ingroup-age-diff}
\end{figure}

Figure~\ref{fig:ingroup-age-diff} illustrates the asymmetric age ingroup effects. We select commenter ages corresponding to the 10\%, 50\%, and 90\% quartiles of the distribution, resulting in ages of 20, 26, and 38. We find that as the commenter gets older, the effect size becomes smaller and statistically non-significant. This is possibly due to older adults having greater familiarity with outgroups when they are asked to make judgments, as studied in \citet{chasteen-2005-seeing-eye}.

\subsection{RQ2: Ingroup Effects in Expressed Text-Based Empathy}

To explore commenters' language use going beyond verdicts, we use LLM-predicted text-based empathy dimensions as proxies to further understand the contextual features of their comments.
Here, we aim to test the set of hypotheses examining whether commenters' language shows more empathy towards the poster, regardless of the social judgments. Similar to RQ1, if ingroup effects exist, we expect to observe positive estimates.

\paragraph{Averaged gender ingroup effects.} Our hypotheses regarding text-based empathy are supported, as illustrated in Figure~\ref{fig:ingroup-overall}.
Across both dimensions, we observe that commenters are significantly more likely to express warmth or compassion and communicate understanding in their comments when interacting with a poster who shares the same gender identity.
Specifically, regarding emotional reactions, the model yields an average positive estimate of 2.38 \textit{pp} (95\% CI: [1.48, 3.28]; $p<0.001$). Similarly, for interpretations, we observe a significant increase of 1.66 \textit{pp} (95\% CI: [0.92, 2.40]; $p<0.001$).
These results confirm that shared gender not only influences the final verdict but actively shapes the empathetic quality of the text itself.

\paragraph{Averaged age ingroup effects.} Our hypotheses on age ingroup effects are also supported: for emotional reactions, the estimate is 1.16 \textit{pp} (95\% CI: [0.41, 1.90]; $p=0.00239$), and for interpretations, the estimate is 0.88 \textit{pp} (95\% CI: [0.25, 1.50]; $p=0.00632$).
In detail, this implies that as the age ratio narrows by a factor of 1.5, the likelihood of commenters expressing emotional reactions increases by 1.15 percentage points, and the likelihood of communicating understanding increases by 0.86 percentage points.

For the asymmetry in age ingroup effects (Figure~\ref{fig:ingroup-age-diff}), we observe the opposite trend in text-based empathy dimensions compared to social judgments. When commenters are older, they are more likely to show ingroup support towards posters of a similar age by expressing understanding in their comments.
This potentially suggests the complexity of how people show understanding and support toward others: commenters may vary in how judgmental they are (the verdict), which interacts with how they choose to phrase their words to the poster (the language).

\subsection{RQ3: Ingroup Effects Moderated by the Social Context}

\begin{figure}[t]
\centering
\includegraphics[width=\linewidth]{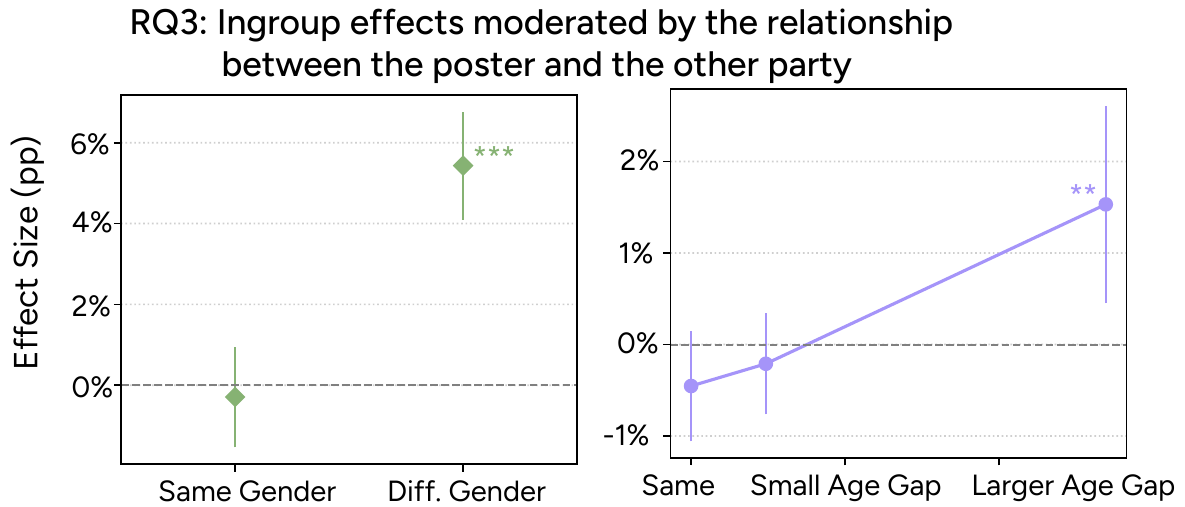}
\caption{Gender and age ingroup effects on social judgments (NTA/YTA) moderated by the relationship between the poster and the other party. When the poster and the other party are of different genders or have a large age gap, the observed average ingroup effects, modeled between the commenter and poster, are larger and significant. 95\% confidence intervals are reported. Significance: ***$p < 0.001$, **$p < 0.01$, *$p < 0.05$.}
\label{fig:ingroup-other}
\end{figure}

To investigate where the ingroup effects are from, we look deeper into the AITA posts. Because these posts describe interpersonal conflicts, the narratives usually present both the poster and the other party with whom they are in conflict. The full set of community verdicts includes combinations of whether the poster or the other party is blamed.

\paragraph{Data subset.}
To this end, we hope to explore the intergroup relationships between the commenter, the poster, and the other party, where a commenter may be ingroup with neither or either. We hypothesize that ingroup effects are stronger when the commenter is ingroup with the poster but outgroup with the other party (i.e., taking sides).
For example (Figure~\ref{fig:example}), a female commenter is ingroup with a female poster but outgroup with the other party (the male fiancé).
We extract the other party’s gender and age using \texttt{gpt-4o-mini-2024-07-18} because regex is difficult to apply in these cases. We gather this information during topic categorization, as identifying the main party first improves task performance. To ensure the LLM does not guess info based on stereotypes, we only use the subset where both gender and age are explicitly extracted. 
\yhadd{On 100 randomly sampled posts, we find only 4 inaccurate predictions, with error patterns similar to those in the poster gender and age extraction task (e.g., predicting or guessing information from circumstantial clues).}
This results in a subset of \underline{71,887} comments on \underline{20,464} posts.\footnote{\yhadd{We get rid of the NAH or ESH comments for this analysis.}}

\paragraph{Mixed-effects model.}
We fit a separate mixed-effects model where the other party's gender ($O_G$) and age ($O_A$) are measured in the same manner as the poster’s, with the social judgment ($V$) as the outcome.
We include two new variables that interact with $\text{IN}_G$ (ingroup gender) and $\text{IN}_A$ (ingroup age):
(1) $\text{OUT}_{\text{PG}}$ is a binary variable indicating whether the other party's gender differs from the poster's: $\text{OUT}_{\text{PG}} = 1$ when they are different, and $\text{OUT}_{\text{PG}} = 0$ indicates the conflict in the post occurs between members of the same gender group;
(2) $\text{OUT}_{\text{PA}}$ represents the age distance between the other party and the poster, calculated as $\text{OUT}_{\text{PA}}= \frac{|P_A - O_A|}{\ln (1.5)}$.
The full formula is:
\begin{equation}
    \begin{aligned}
        \nonumber \text{logit}(P(V_i = 1)) &= \beta_0 + \beta_1 \text{IN}_{Gi} + \beta_2 \text{OUT}_{\text{PG}i} \\
        &\quad + \beta_3 (\text{IN}_{Gi} \times \text{OUT}_{\text{PG}i}) \\
        &\quad + \beta_4 \text{IN}_{Ai} + \beta_5 \text{OUT}_{\text{PA}i} \\
        &\quad + \beta_6 (\text{IN}_{Ai} \times \text{OUT}_{\text{PA}i}) \\
        &\quad + \beta_7 C_{Ai} + \beta_8 (\text{IN}_{Ai} \times C_{Ai}) \\
        &\quad + \boldsymbol{\gamma}^T \mathbf{X}'i  + u_{P_{\text{ID}}[i]} + v_{C_{\text{ID}}[i]}
    \end{aligned}
\end{equation}
In addition to the existing ones, $\mathbf{X}'_i$ includes $O_G$ and $O_A$.

\paragraph{Averaged gender ingroup effects.} On average, we still find a significant positive gender ingroup estimate. This means that regardless of the gender of the other party, when the commenter is ingroup with the poster, commenters are 3.96 \textit{pp} (95\% CI: [3.09, 4.82]; $p<0.001$) more likely to issue an NTA verdict. However, these effects are moderated by the relationship between the poster and the other party (Figure~\ref{fig:ingroup-other}).
When the other party and the poster are from different gender groups, the estimate for $\text{IN}_G$ is much larger and significant at 5.43 \textit{pp} (95\% CI: [4.10, 6.77]; $p<0.001$). In contrast, when the other party and the poster are of the same gender, the effect is non-significant at -0.29 \textit{pp} (95\% CI: [-1.54, 0.95]; $p=0.645$).

\paragraph{Averaged age ingroup effects.} Regarding the age ingroup, the average estimate for $\text{IN}_A$ is not significant under the new setup, at 0.38 \textit{pp} (95\% CI: [-0.63, 0.88]; $p=0.746$). However, the results (Figure~\ref{fig:ingroup-other})---where we sample the 10\%, 50\%, and 90\% quartiles of the $\text{OUT}_{\text{PA}}$ distribution to represent a negligible, small, and large age gap between the poster and the other party---show a different pattern. 
When the poster and the other party have a larger age gap ($\text{OUT}_{\text{PA}} = 1.348$), the estimate for $\text{IN}_A$ becomes positive and significant at 1.53 \textit{pp} (95\% CI: [0.45, 2.61]; $p=0.00543$).
This suggests that, similar to the moderation effects of gender, ingroup effects are stronger when the commenter is an age-ingroup member with the poster but not with the other party.

Overall, results indicate that intergroup bias correlates with the expressions of social judgments and empathy in the comments of the AITA community. However, this bias is dependent on the social context and group nuances.


\section{Discussion}
\label{sec:discussion}

In this work, we study intergroup bias based on gender and age using online social interactions in the r/AmITheAsshole community. We address three research questions, building on a final dataset of 126,716 comments and 36,163 posts.
We focus on a specific type of intergroup behavior where a commenter reads a post with self-reported gender and age and writes a comment containing a clear verdict and open-ended words.
Because these posts describe real-world interpersonal conflicts, the comments show how commenters understand the posters based on their stories.
First, we look into social judgments where commenters decide if the poster is to blame.
Then, we extend the analysis to their open-ended ``rationales'' and develop text-based empathy scales specific to those comments.
Finally, we revisit the posts to understand in depth how this intergroup process happens, specifically regarding making social judgments.
After carefully controlling for effects from posts, posters, and commenters, we find that when the commenter and poster share the same gender or are of a similar age, commenters are more likely to issue an NTA verdict and show greater understanding through the open-ended comments. In addition, the age ingroup effects are not the same among older and younger adults, and these ingroup effects are moderated by the intergroup relationships in the post.
To conclude, understanding others and expressing this understanding only appears to be straightforward. Social group identities, like gender and age, are two of many perspectives for interpreting this human behavior.
In addition, there are other shared experiences we have with others that influence our cognitive empathy, such as being on the same sports team, speaking the same language, or growing up in the same town.
In a future where AI increasingly adopts human-like roles---whether as a poster mimicking various text styles~\cite{basoah-etal-2025-not-like-us}, a commenter pretending to behave like a human~\cite{cheng2025sycophanticaidecreasesprosocial}, or a communication mediator~\cite{jakesch-etal-2019-ai-mediated}---will it bridge the gap of shared experiences, or will it complicate the intergroup biases? We leave this for future work.

\paragraph{Implications.} \yhadd{We report the observed effects of other predictors in the regression model for RQ1. Post topic categories show average marginal effects (AMEs) ranging from $-5.56$ \textit{pp} (romance) to $-1.55$ \textit{pp} (work/school) relative to the public category. In comparison, the gender ingroup effect has an AME of $3.18$ \textit{pp}, and a one-unit increase in log age distance corresponds to an AME of $1.03$ \textit{pp}. Because social judgment is a relatively indirect outcome of the intergroup empathy gap, other factors also influence the outcome. Between the two ingroup effects, the gender ingroup effect is larger than the age ingroup effect.}

\paragraph{Limitations and future work.} Our work has a number of limitations that can be further studied.
First, from the initial dataset, we apply many rules to ensure a relatively accurate user database from which to build intergroup relationships, resulting in a large portion of the data being discarded. Even though we do not observe a strong difference between self-disclosed users and other users, because we omit any poster--commenter pairs where the identity of either is unknown, this still risks leading to potential selection bias where our results only represent user interactions in which both identities are known and expressed explicitly in ways we can parse. We prioritize correct user gender and age labels using both regex and LLM approaches to avoid the potential ethical harms or invalidity of inferring gender or age from indirect evidence. Future work can refine extraction rules or methods to infer user identities in online communities.

Second, we would like to discuss the distinction between self-identity and perceived identity. For the intergroup effects in our study, we care about commenters' self-identity (which is based on their self-reported identity when they take the poster role) and how the commenter perceives the identity of the poster and the other parties in the poster's stories.
Because we cannot know exactly how those posts are received or with what kinds of perceived group identities, we only study posts with clearly self-disclosed identities and assume those are consistent with the post content and furthermore received the same way by the commenter. It is possible that people do not share their real identities online, that their self-disclosed identities might not match the post content, or that commenters view the posters based on other interactions. We hope future work can address these limitations and study intergroup interactions under more complex scenarios.

Third, we acknowledge the limitations of using LLMs as data annotators~\cite{egami-etal-2023-using-imperfect, gligoric-etal-2025-unconfident}, though we try to carefully examine the data. 
We consider using LLMs as a convenient and powerful way to extend RQ1 with a deeper understanding of both the comment and post content, compared to using only the extracted social identities and judgments, for exploratory purposes. 
Future work could build more robust annotation pipelines or approaches and extend the current study setup.

\yhadd{Lastly, our final data range from 2018 to 2022. This was the data we could find online at the time, and we do not intend to exclude data from post-pandemic years. While intergroup bias is well studied, we understand that discourse norms are changing, which may influence the findings. Future work can apply our approach to more recent data and examine how the findings may have shifted.}




\paragraph{Ethical concerns.} Though we only use publicly available data with user-reported gender and age to build intergroup relationships, it is important to note that Reddit users have not consented to research studies. While r/AmITheAsshole data has been used widely in previous research, we will only release anonymous, aggregated features for this analysis to protect user privacy.






\section*{Acknowledgments}

We would like to thank the anonymous reviewers for their valuable feedback. 
This material is based upon work partially supported by the NSF under Grant No. 2229885 (NSF Institute for Trustworthy AI in Law and Society, TRAILS), and NSF CAREER Award No. 2339746 (Rudinger).
Any opinions, findings and conclusions or recommendations expressed in this material are those of the author(s) and do not necessarily reflect the views of the National Science Foundation.

\bibliography{aaai2026}

\section*{Paper Checklist}

\begin{enumerate}

\item For most authors...
\begin{enumerate}
    \item  Would answering this research question advance science without violating social contracts, such as violating privacy norms, perpetuating unfair profiling, exacerbating the socio-economic divide, or implying disrespect to societies or cultures?
    \answerYes{Yes, discussed in \S\ref{sec:intro} and \S\ref{sec:discussion}.}
  \item Do your main claims in the abstract and introduction accurately reflect the paper's contributions and scope?
    \answerYes{Yes}
   \item Do you clarify how the proposed methodological approach is appropriate for the claims made? 
    \answerYes{Yes, discussed in \S\ref{sec:background}.}
   \item Do you clarify what are possible artifacts in the data used, given population-specific distributions?
    \answerYes{Yes, discussed in \S\ref{sec:data}.}
  \item Did you describe the limitations of your work?
    \answerYes{Yes, discussed in \S\ref{sec:discussion}.}
  \item Did you discuss any potential negative societal impacts of your work?
    \answerYes{Yes, discussed in \S\ref{sec:discussion}.}
      \item Did you discuss any potential misuse of your work?
    \answerYes{Yes, discussed in \S\ref{sec:discussion}.}
    \item Did you describe steps taken to prevent or mitigate potential negative outcomes of the research, such as data and model documentation, data anonymization, responsible release, access control, and the reproducibility of findings?
    \answerYes{Yes, discussed in \S\ref{sec:discussion}.}
  \item Have you read the ethics review guidelines and ensured that your paper conforms to them?
    \answerYes{Yes}
\end{enumerate}

\item Additionally, if your study involves hypotheses testing...
\begin{enumerate}
  \item Did you clearly state the assumptions underlying all theoretical results?
    \answerYes{Yes, discussed in \S\ref{sec:intro} and \S\ref{sec:results}.}
  \item Have you provided justifications for all theoretical results?
    \answerYes{Yes, discussed in \S\ref{sec:intro}, \S\ref{sec:data} and \S\ref{sec:results}.}
  \item Did you discuss competing hypotheses or theories that might challenge or complement your theoretical results?
    \answerYes{Yes, discussed in \S\ref{sec:results}.}
  \item Have you considered alternative mechanisms or explanations that might account for the same outcomes observed in your study?
    \answerYes{Yes, discussed in \S\ref{sec:results}.}
  \item Did you address potential biases or limitations in your theoretical framework?
    \answerYes{Yes, discussed in \S\ref{sec:discussion}.}
  \item Have you related your theoretical results to the existing literature in social science?
    \answerYes{Yes, discussed in \S\ref{sec:results}.}
  \item Did you discuss the implications of your theoretical results for policy, practice, or further research in the social science domain?
    \answerYes{Yes, discussed in \S\ref{sec:discussion}.}
\end{enumerate}

\item Additionally, if you are including theoretical proofs...
\begin{enumerate}
  \item Did you state the full set of assumptions of all theoretical results?
    \answerNA{Not Applicable.}
	\item Did you include complete proofs of all theoretical results?
    \answerNA{Not Applicable.}
\end{enumerate}

\item Additionally, if you ran machine learning experiments...
\begin{enumerate}
  \item Did you include the code, data, and instructions needed to reproduce the main experimental results (either in the supplemental material or as a URL)?
    \answerYes{Yes, discussed in \S\ref{sec:data} and \S\ref{app:llm-prompt}. Code and data will be released upon publication.}
  \item Did you specify all the training details (e.g., data splits, hyperparameters, how they were chosen)?
    \answerYes{Yes, discussed in \S\ref{sec:data} and \S\ref{app:llm-prompt}.}
     \item Did you report error bars (e.g., with respect to the random seed after running experiments multiple times)?
    \answerYes{Yes, discussed in \S\ref{sec:results}.}
	\item Did you include the total amount of compute and the type of resources used (e.g., type of GPUs, internal cluster, or cloud provider)?
     \answerYes{Yes, discussed in \S\ref{sec:data}.}
     \item Do you justify how the proposed evaluation is sufficient and appropriate to the claims made? 
     \answerYes{Yes, discussed in \S\ref{sec:data} and \S\ref{app:llm-prompt}.}
     \item Do you discuss what is ``the cost`` of misclassification and fault (in)tolerance?
     \answerYes{Yes, discussed in \S\ref{sec:data}, \S\ref{sec:discussion} and \S\ref{app:llm-prompt}.}
  
\end{enumerate}

\item Additionally, if you are using existing assets (e.g., code, data, models) or curating/releasing new assets, \textbf{without compromising anonymity}...
\begin{enumerate}
  \item If your work uses existing assets, did you cite the creators?
    \answerYes{Yes, discussed in \S\ref{sec:data}}
  \item Did you mention the license of the assets?
    \answerYes{Yes, discussed in \S\ref{sec:data}}
  \item Did you include any new assets in the supplemental material or as a URL?
    \answerNA{Not Applicable. No new assets.}
  \item Did you discuss whether and how consent was obtained from people whose data you're using/curating?
    \answerYes{Yes, discussed in \S\ref{sec:discussion}; but also NA as we use public Reddit data.}
  \item Did you discuss whether the data you are using/curating contains personally identifiable information or offensive content?
    \answerYes{Yes, discussed in \S\ref{sec:data} and \S\ref{sec:discussion}.}
\item If you are curating or releasing new datasets, did you discuss how you intend to make your datasets FAIR (see \citet{fair})?
\answerNA{Not Applicable.}
\item If you are curating or releasing new datasets, did you create a Datasheet for the Dataset (see \citet{gebru2021datasheets})? 
\answerNA{Not Applicable.}
\end{enumerate}

\item Additionally, if you used crowdsourcing or conducted research with human subjects, \textbf{without compromising anonymity}...
\begin{enumerate}
  \item Did you include the full text of instructions given to participants and screenshots?
    \answerNA{Not Applicable.}
  \item Did you describe any potential participant risks, with mentions of Institutional Review Board (IRB) approvals?
    \answerNA{Not Applicable.}
  \item Did you include the estimated hourly wage paid to participants and the total amount spent on participant compensation?
    \answerNA{Not Applicable.}
   \item Did you discuss how data is stored, shared, and deidentified?
   \answerNA{Not Applicable.}
\end{enumerate}

\end{enumerate}

\appendix
\section{LLM Prompting Experiment Details}
\label{app:llm-prompt}

We present the detailed prompts used in this study.

For Prompt A.1 on the user information extraction task, because our goal is to use the LLM to clean up noise while relying on regex extraction results, LLM errors have very little influence on our analysis.
For each of the other prompts, we develop them iteratively with an annotated subset of 24 comments and 11 posts among the authors. Note that the number is relatively small because the goal is to go through these examples one by one, analyze the error patterns, and develop and update the prompts accordingly.
To avoid unreliable predictions impacting our findings, we consider these only as supplementary analyses (RQ2: understanding contextual language features; RQ3: investigating cases where the other party's information is presented) to enhance the understanding of intergroup bias broadly.

\subsection{Prompt A.1: Post User Info Extraction System Prompt}
\label{prompt:post-user}
\begin{MyVerbatim}
You are an expert in extracting information from Reddit posts.
You will be given a post (including title and content) and you need to extract ONLY the information that is explicitly mentioned within the post text itself. Do not make assumptions or inferences. Ignore external links in the post.

The information you need to extract is:
- The gender of the poster: Look for explicit mentions such as "M", "F", "male", "female", or clear self-identifying statements within the post title or content
- The age of the poster: Look for explicit age mentions (1-100) within the post title or content, such as "I am 25" or "25M" or similar direct statements

Important guidelines:
- Extract information ONLY if it is directly stated in the post title or content
- Do not infer or assume information based on context, usernames, or indirect references
- Do not use external knowledge or stereotypes

Output the information in JSON format with the keys "gender" and "age".
If any information is not explicitly mentioned in the post, return "unknown" as the value for that key.
\end{MyVerbatim}

\subsection{Prompt A.2: Post Conflict Party \& Topic System Prompt}
\begin{MyVerbatim}
You are an expert at analyzing Reddit posts.
You will be given a post (including title and content) and identify the main conflict, then classify it into one of the following categories.

Please follow the following steps to analyze the post:
1. Identify the OP (Original Poster): the OP is the person writing the post asking "Am I the Asshole"
2. Identify the main conflict person: find the PRIMARY person the OP is in conflict with. This person is usually mentioned in the title or early in the post. If multiple people are involved, identify who is MOST CENTRAL to the conflict. Ask yourself: "Who is OP primarily asking if they were an asshole toward?"
3. Extract the main person's gender and age (if only there is ONE clear main person, otherwise return "unknown"): look for explicitly stated age and gender (e.g., "29F", "my brother (35M)", "my wife 42F"). Extract information ONLY if it is directly stated. Do not infer or assume information based on context, usernames, or indirect references.
4. Classify the relationship category: the category is based on WHO the main person IS to the OP, NOT what the conflict is about. If the main person is not clear, judge the category based on the main person(s) and the conflict itself.
Examples:
- If OP's friend did something involving the friend's sister, and OP is in conflict with the friend, choose [3] Friendship (not Family)
- If OP's spouse's coworker is rude, and OP is in conflict with the coworker, choose [5] Public & Strangers (not Romance)

Relationship Categories (ordered by priority):
[1] Romance: the main person is OP's romantic partner, dating partner, spouse, ex-partner, or crush
[2] Family: the main person is OP's relative (parent, sibling, child, in-law, grandparent, aunt, uncle, cousin, etc.)
[3] Friendship: the main person is OP's friend or best friend
[4] Work/School: the main person is OP's colleague, boss, employee, teacher, student, or classmate
[5] Public & Strangers: the main person is OP's neighbor, landlord, tenant, service worker, or a stranger in public (store, transport, restaurant, etc.)
[6] Society: the conflict is about broad societal/ethical issues (politics, systemic racism, gender debates) rather than a specific person. This category is the lowest priority.

Important guidelines:
- Do not use external knowledge or stereotypes
- Ignore external links and images
- If unclear between two categories, use the HIGHER priority one (lower number = higher priority)

Output the information in JSON format with the keys "main_person", "age", "gender", and "category".
Provide your answer as valid JSON only, with no additional text.
```json
{
  "main_person": "brief description of who the main person is to OP if clear, otherwise try to describe the main person(s) and the conflict itself",
  "age": "number or unknown",
  "gender": "M or F or unknown",
  "category": number (1-6) // 1-6 are the relationship categories
}
```
\end{MyVerbatim}

\subsection{Prompt A.3: Post Input Prompt (Both Experiments)}
\begin{MyVerbatim}
POST: {post}

OUTPUT:
```json
\end{MyVerbatim}

\subsection{Prompt A.4: Comment Text-Based Empathy System Prompt}
\begin{MyVerbatim}
You are an expert at analyzing r/AmITheAsshole subreddit comments.
You will be given a post written by the Original Poster (OP) and a comment responding to that post. Your task is to assess the empathy expressed in the comment SPECIFICALLY TOWARDS THE OP (the person who wrote the post) along two dimensions:

Dimension 1: Emotional Reactions
Does the comment express or allude to warmth, compassion, concern, or similar feelings of the commenter towards the OP?

0 - Comment not expressing or alluding to emotional reactions towards the OP at all
1 - Comment alluding to emotional reactions towards the OP but not expressing them explicitly
2 - Comment expressing emotional reactions towards the OP explicitly

Dimension 2: Interpretations
Does the comment communicate an understanding of the OP's experiences and feelings?

0 - Comment not communicating any understanding of the OP's experiences and feelings at all
1 - Comment communicating only generic understanding of the OP's situation
2 - Comment communicating specific understanding of the OP, including at least one of these signals:
   - Contains conjectures or speculations about the OP's experiences and/or feelings
   - Contains paraphrases of the OP's experiences and/or feelings
   - Reflects on similar experiences of their own or others that relate to the OP's situation
   - Describes similar experiences of their own or others that relate to the OP's situation

Important guidelines:
- Think step by step. First, identify who the OP is. Then, assess the empathy directed towards the OP.
- For example:
* Post (summary): A 20-year-old woman describes how her family repeatedly criticized her 14-year-old sister's weight despite the sister having a healthy BMI. The family called her names like "cow" and "whale" during arguments, and the OP suggested her sister should fast. The younger sister subsequently developed what appears to be a severe eating disorder, becoming skeletal over 8 months and refusing to eat. The OP now wonders if she's TA and worries her words may have contributed to her sister's condition.
* Comment: "You guys pushed that poor girl into an eating disorder. Yes YTA and you should be ashamed"
* Explanation: The comment expresses strong emotion (anger, sympathy for the sister, condemnation), but these emotions are directed AT the OP negatively and TOWARD the younger sister protectively. There is no warmth, compassion, or concern expressed toward the OP herself, and no demonstration of understanding the OP's current worry, guilt, or internal conflict. Therefore annotations should be 0 for both emotional reactions and interpretations.

Output the information in JSON format with the keys "emotional_reactions", "interpretations".
Provide your answer as valid JSON only, with no additional text.
```json
{
  "emotional_reactions": number (0-2),
  "interpretations": number (0-2)
}
```
\end{MyVerbatim}

\subsection{Prompt A.5: Comment Input Prompt}
\begin{MyVerbatim}
POST: {post}
COMMENT: {comment}

OUTPUT:
```json
\end{MyVerbatim}

\end{document}